%% file: root.tex
\documentclass[letterpaper, 10 pt, conference]{ieeeconf}  

\IEEEoverridecommandlockouts                              

\usepackage{graphicx}
\usepackage{amsmath} 
\usepackage{hyperref}
\usepackage{bm}
\usepackage{multirow}
\usepackage{comment}
\usepackage{xcolor}
\usepackage{tikz}

\title{\LARGE \bf
Multi-Camera Hand-Eye Calibration for Human-Robot Collaboration in Industrial Robotic Workcells 
}

\author{Davide Allegro$^{1}$, Matteo Terreran$^{1}$ and Stefano Ghidoni$^{1}$
\thanks{$^{1}$All the authors are with the Department
of Information Engineering (DEI) at the University of Padova, via Gradenigo
6/B, 35131 Padova, Italy. 
        {\tt\small Email: davide.allegro.1@phd.unipd.it, [matteo.terreran; stefano.ghidoni]@unipd.it}}%
}

\begin{document}

\maketitle
\thispagestyle{empty}
\pagestyle{empty}


\newcommand\submittedtext{%
  \footnotesize This work has been submitted to the IEEE for possible publication. Copyright may be transferred without notice, after which this version may no longer be accessible.}

\newcommand\submittednotice{%
\begin{tikzpicture}[remember picture,overlay]
\node[anchor=south,yshift=10pt] at (current page.south) {\fbox{\parbox{\dimexpr0.65\textwidth-\fboxsep-\fboxrule\relax}{\submittedtext}}};
\end{tikzpicture}%
}

\begin{abstract}

In industrial scenarios, effective human-robot collaboration relies on multi-camera systems to robustly monitor human operators despite the occlusions that typically show up in a robotic workcell.
In this scenario, precise localization of the person in the robot coordinate system is essential, making the hand-eye calibration of the camera network critical. 
This process presents significant challenges when high calibration accuracy should be achieved in short time to minimize production downtime, and when dealing with extensive camera networks used for monitoring wide areas, such as industrial robotic workcells.
Our paper introduces an innovative and robust multi-camera hand-eye calibration method, designed to optimize each camera’s pose relative to both the robot’s base and to each other camera. This optimization integrates two types of key constraints: i) a single board-to-end-effector transformation, and ii) the relative camera-to-camera transformations. We demonstrate the superior performance of our method through comprehensive experiments employing the METRIC dataset and real-world data collected on industrial scenarios, showing notable advancements over state-of-the-art techniques even using less than 10 images. Additionally, we release an open-source version of our multi-camera hand-eye calibration algorithm at \url{https://github.com/davidea97/Multi-Camera-Hand-Eye-Calibration.git}.
\end{abstract}

\section{INTRODUCTION}
\label{sec:introduction}
\submittednotice
Human-robot collaboration (HRC) aims to a close and direct interaction between humans and robots to achieve a common objective, leveraging the synergy between human intelligence and manipulation capabilities and robot precision~\cite{ajoudani2018progress, kim2021human,lorenzini2023ergonomic}. 
This collaborative pattern is spreading significantly in industries, fostering greater production flexibility while maintaining efficiency and productivity~\cite{villani2018survey, matheson2019human, simoes2022designing}.
\begin{figure}[t]
  \centering
  \includegraphics[width=0.49\textwidth]{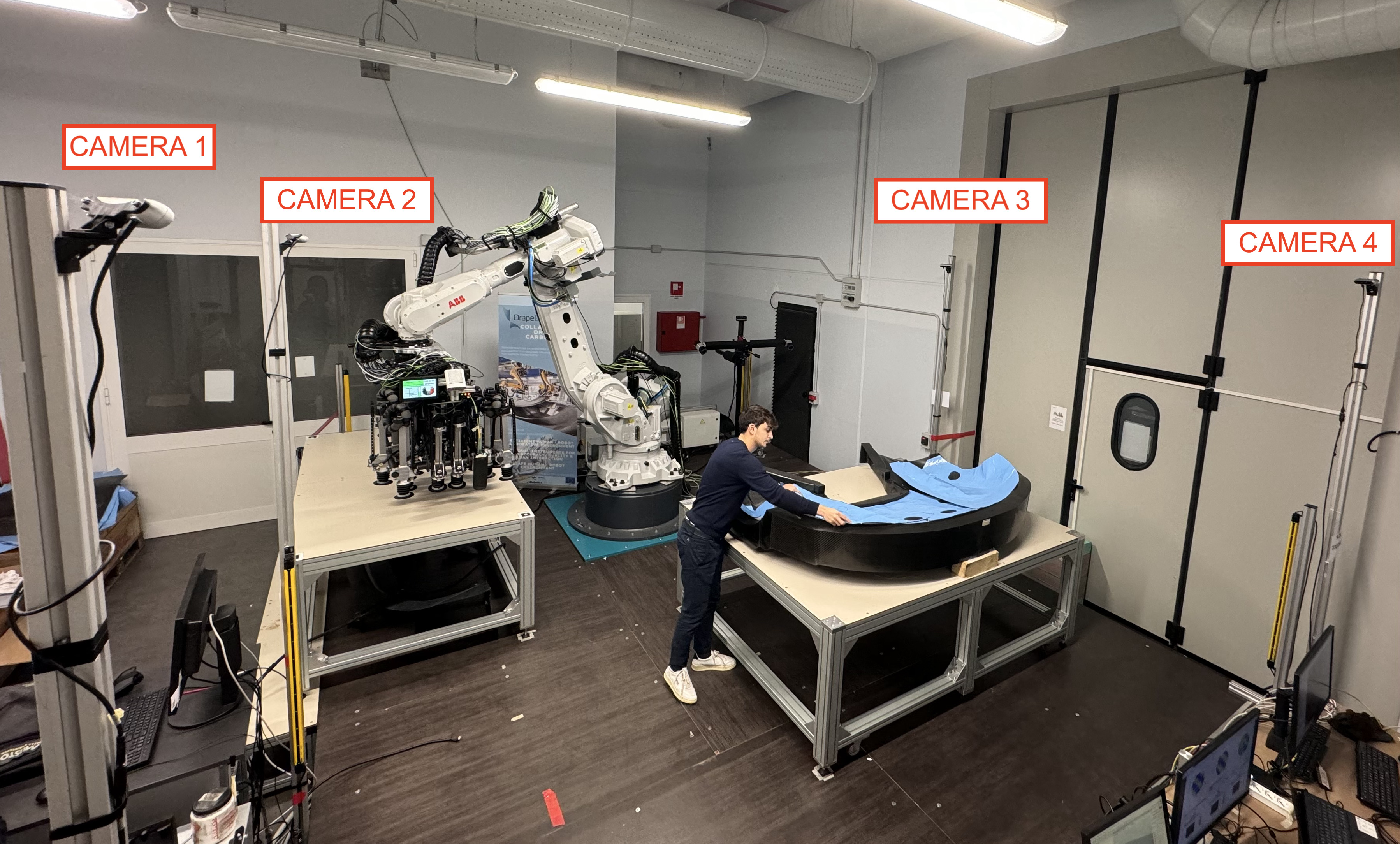}
  \caption{A large industrial robotic workcell equipped with a camera network around an ABB robot arm, enabling human-robot collaboration task in a carbon fiber draping process as foreseen in the DrapeBot European Project.}
  \label{fig:robotic_workcell}
  \vspace{-0.5cm}
\end{figure}
Several projects dealing with industrial scenarios, such as Sharework\footnote{https://sharework-project.eu/} and DrapeBot\footnote{https://www.drapebot.eu/}, have recently proposed to supervise the robotic workcell avoiding occlusions problems by means of a multi-camera system positioned around a robot arm, as shown in Figure~\ref{fig:robotic_workcell}. This enables the continuous monitoring of both the robot workspace and human worker activities throughout the collaboration process~\cite{orsag2023towards, terreran2020low}. 
As the robot and each sensor natively defines its own reference system, it is essential to express information provided by each sensor in a common reference frame---a convenient option here is the robot base coordinates system, to make the robot aware of its surroundings.
Such process is generally known as hand-eye calibration, and aims to determine the relative transformation between the robot base and a camera by moving a calibration pattern attached to the robot end-effector to different positions in front of the camera~\cite{su2022research}.
%
%

When dealing with multiple cameras, calibrating all of them with respect to the robot can be a challenging task: 
i) the calibration pattern must be compact to avoid collisions during the robot’s movement; ii) the calibration often involves a large camera network, necessary to monitor the whole robotic workcell, dealing with rather large distances among cameras and robot; iii) the calibration process needed to be performed in short time to reduce as much as possible production line downtime; iv) accurate calibration must be provided even with a limited number of images, which is a common occurrence in industrial environments due to the difficulty of moving the robot arm safely in a cluttered space.

In existing literature, only a few works have addressed the challenge of hand-eye calibration for camera networks, particularly in industrial scenarios~\cite{tabb2017solving}. When dealing with multiple cameras, traditional methods typically perform hand-eye calibration of each camera separately~\cite{miseikis2016automatic}. This process finds the optimal transformation of each sensor with respect to the robot, then derives relative transformations among cameras either by linking transformations together or through additional stereo calibrations between the camera pairs~\cite{rashd2020open}. This approach often leads to methods that are neither robust nor precise, especially in demanding scenarios like industrial ones, where the risk for error propagation to the final calibration is significantly high~\cite{wang2022accurate}.
Moreover, these existing methods usually focus on relatively small robotic workcells, positioning cameras approximately 1 meter away from both the calibration pattern and each other~\cite{tabb2019calibration}. This can be considered a significant limitation creating a gap between the capabilities of current calibration techniques and the demands of industrial environments.

In this paper we introduce a non-linear optimization algorithm to address hand-eye calibration in multi-camera setups within industrial robotic workcells. 
Our method generalizes the work presented in \cite{evangelista2022unified} to multi-camera systems, enabling the simultaneous pose estimation of each camera with respect to all other sensors and to the robot's base reference frame. 
Unlike conventional methods, which usually focus on calibrating each camera independently, our method 
introduces two main types of constraint to better optimize the mutual pose of the cameras: i) a single board-to-end-effector transformation and ii) the relative rototranslations camera-to-camera.
The former allows to streamline the process eliminating the redundancy of determining that transformation for each individual hand-eye calibration; the latter ensures the optimization of the relative transformation among cameras by exploiting the simultaneous detection of multiple cameras of the calibration pattern.
This approach guarantees consistency across poses of all cameras and enhances calibration performance by preventing error propagation that can occur with individual calibrations and the need for further calibration steps to determine transformations between cameras. 
Generally, when considering camera networks for monitoring robotic workcells, cameras are strategically placed to reduce occlusions and simultaneously capture different viewpoints. This setup easily leads to the simultaneous acquisition of images of the same calibration pattern during calibration, allowing our method to concurrently leverage multi-camera information without imposing stringent design requirements on the workcell.

Extensive evaluations on the synthetic and real data of the open source METRIC\footnote{https://zenodo.org/records/7976757} dataset~\cite{allegro2023metric} allowed to investigate the impact of the workcell and pattern sizes on the calibration performances. Additionally, comprehensive experiments on real industrial robotic workcells were necessary to validate the proposed method's robustness, precision, and applicability in demanding industrial environments.

In summary, our work offers three main contributions:
\begin{enumerate}
    \item A novel multi-camera hand-eye calibration method for calibrating multiple sensors with respect to a robot and to each others, characterized by two key constraints in the optimization procedure: a single board-to-end-effector and relative camera-to-camera transformations;
    \item A comprehensive performance evaluation of the proposed method against state-of-the-art hand-eye calibration techniques using the METRIC dataset;
    \item A thorough comparative analysis of our approach in real-world industrial settings, outperforming state-of-the-art methods in  challenging scenarios of large camera network and limited number of images available for each camera.
\end{enumerate}


\section{RELATED WORKS}
\label{sec:related_works}

\begin{figure}[t]
  \centering
  \includegraphics[width=1\columnwidth]{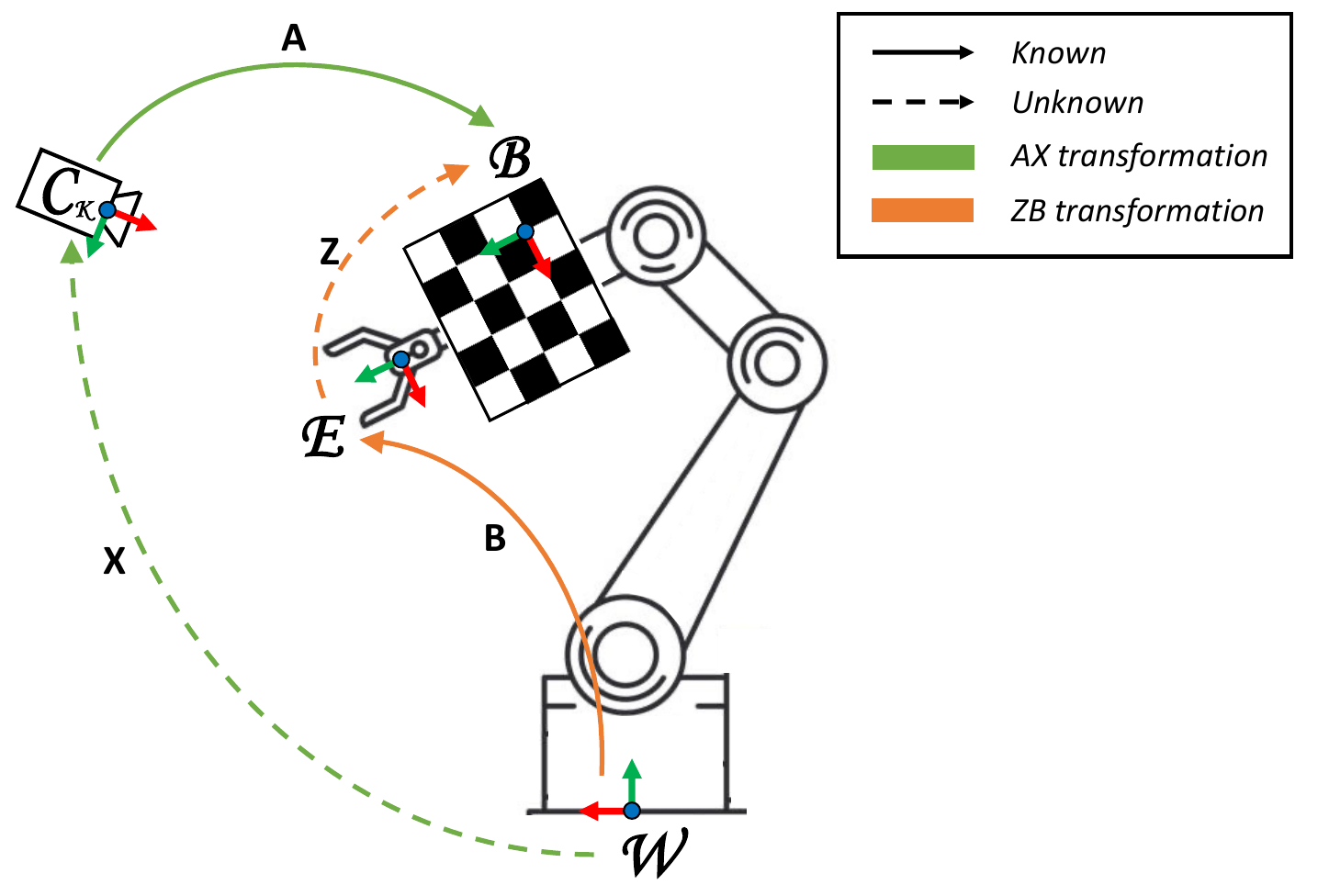}
  \caption{Single-camera hand-eye calibration setup, used to derive the homogeneous transformations $AX=ZB$. Here, $A$ denotes the camera-to-board transformation and $B$ represents the pose of the robot's end-effector relative to its base $W$. While $X$ and $Z$ are the two unknown transformations.}
  \label{fig:hand_eye_setup}
  \vspace{-0.5cm}
\end{figure}

\textbf{Single-camera hand-eye calibration.} 
In the literature, several methods have been proposed to tackle hand-eye calibration in single-camera setups. Some of these approaches evaluate the solutions to the homogeneous equation $AX=ZB$ as shown in Figure~\ref{fig:hand_eye_setup}, with the aim of minimizing translation and rotation errors. 
Here, $A$ and $B$ denote the camera-to-board and the transformation between the robot's end-effector with respect to its base, respectively. While $X$ and $Z$ are the unknown transformations that have to be estimated.
Among these, Tsai \emph{et al.}~\cite{tsai1989new} estimated separately translation and rotation with angle-axis representation,
Park \emph{et al.}~\cite{park1994robot} proposed a solution based on Lie algebra, Daniilidis \emph{et al.} used a dual quaternion parametrization~\cite{daniilidis1996dual}, while Liang \emph{et al.}~\cite{liang2008hand} and Andreff \emph{et al.}~\cite{andreff1999line} proposed the Kronecker product parametrization. More recently, Shah \emph{et al.}~\cite{shah2013solving} formulated a closed-form solution using an SVD-based algorithm and the Kronecker product to solve for rotation and translation separately; Li \emph{et al.}~\cite{li2010simultaneous} employed both Kronecker product and dual quaternions to solve the hand-eye calibration problem.
However, all these methods depend on directly estimating the board-to-camera transformation $A$ using the Perspective-n-Point algorithm~\cite{schweighofer2008globally}. This approach can introduce errors, particularly when handling blurred images that hinder precise pattern detection. 

On the other hand, alternative methods are based on the minimization of visual quantities, specifically the re-projection error.
This process involves minimizing the difference between the observed control points of the calibration pattern attached to the robot end-effector and their corresponding re-projected points, (i.e., the 3D control points of the calibration pattern re-projected back onto the camera's image plane).
In this context, multiple methods have been introduced: Evangelista \emph{et al.} presented an hand-eye calibration method for single-camera configurations across different setups~\cite{evangelista2022unified}. Koide \emph{et al.} proposed an hand-eye calibration method, implementing the minimization of the re-projection error as a pose graph optimization problem, demonstrating high accuracy at a high computational cost~\cite{koide2019general}. These approaches offer a notable advantage by directly leveraging the calibration pattern images, removing the need for explicit camera pose estimation, which typically requires the PnP algorithms~\cite{malti2013hand, tabb2015parameterizations}. 

\textbf{Multi-camera hand-eye calibration.} 
In scenarios involving multiple cameras, the spatial transformation between cameras is often determined either by performing hand-eye calibration for each individual camera and applying a transformation chain, or by calibrating just one camera using hand-eye calibration followed by stereo camera calibration~\cite{rashd2020open}. However, both approaches carry the risk of error propagation.
Only few works in the literature address the simultaneous calibration of a camera network with respect to a robot, probably because of the complexity of the task, which involves managing and integrating visual data from multiple perspectives at the same time. Wang \emph{et al.}~\cite{wang2022accurate} proposed a multi-camera calibration method to handle non-overlapping camera network setups, however relying on an external motion capture system to achieve precise camera position during the calibration process.
Tabb~\cite{tabb2017solving} proposed a robot-world hand-multiple-eye calibration for a small robotic workcell, relying on the minimization of the corner re-projection error, considering the board-to-end-effector transformation $Z$ to be unique for all cameras. 
Evangelista \emph{et al.} \cite{evangelista2023graph} proposed an hand-eye calibration method for a multi-camera setup based on a pose-graph optimization, proving to be accurate, but at the same time really time-consuming. 
\begin{figure}[t]
  \centering
  \includegraphics[width=0.49\textwidth]{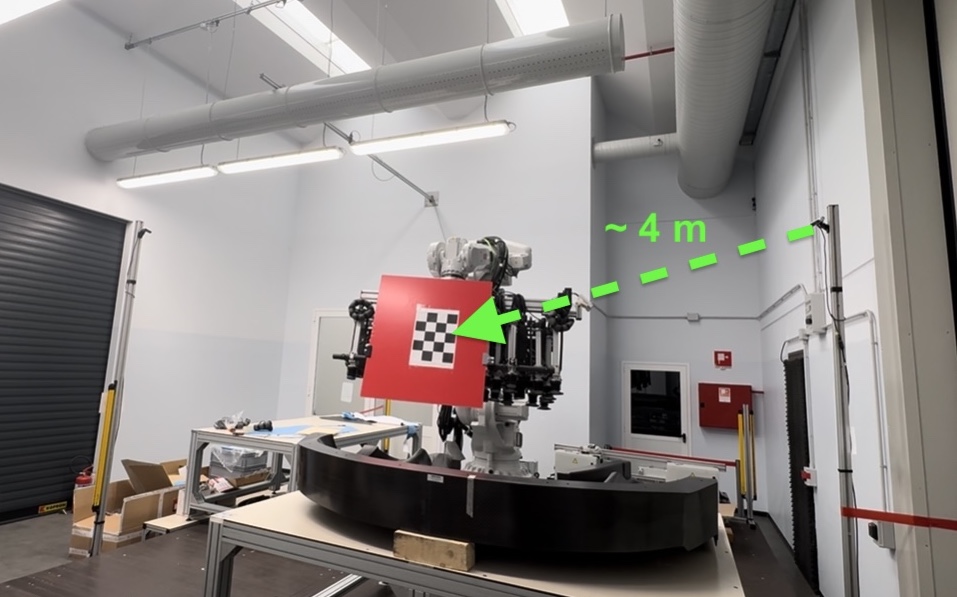}
  \caption{Industrial robotic workcell calibration; a checkerboard attached to the robot end-effector is moved to different positions in front of the surrounding sensors at about 4 meters away from the calibration pattern.}
  \label{fig:workcell_calib}
  \vspace{-0.5cm}
\end{figure}
A notable limitation of these methods is their effectiveness mainly within small robotic workcells.
The underlying assumption of these methods is that the cameras must be positioned relatively close to the calibration pattern, typically within a distance of about one meter, to ensure precise detection of the board. This proximity requirement, however, often does not align with the spatial configurations commonly found in real-world industrial environments. In many practical settings, especially in larger or more complex workcells, the calibration pattern, and thus the robot, needs to be placed further from the cameras to accommodate the operational layout and the movement of humans and robots within the workspace, as depicted in Figure\ref{fig:workcell_calib}. 
In this scenario, the detection of the calibration pattern becomes challenging, negatively affecting the final calibration. Additionally, many of aforementioned approaches require a significant number of images to converge to an optimal solution, which is challenging to ensure in industrial scenarios. This discussion highlights a critical gap between existing calibration methods and the real needs of industrial environments, a gap that is addressed in our work.

\section{METHODOLOGY}
\label{sec:methodology}
%
This section introduces our novel multi-camera hand-eye calibration method, that generalizes the approach introduced in~\cite{evangelista2022unified} to configurations involving multiple cameras. As in~\cite{evangelista2022unified}, our calibration method is based on the minimization of the re-projection error by means of non-linear optimization, but it introduces two main constraints: i) a single transformation between the calibration pattern and the robot's end-effector removing the need for calculating that transformation independently for each camera, and ii) the spatial transformation among the cameras ensuring consistency among all relative transformations between the cameras and the robot base.
Section \ref{subsec:single_calib} briefly summarizes the single-camera hand-eye calibration work proposed in \cite{evangelista2022unified}, presenting the main notations that will be adopted for the formulation of our method. In Section \ref{subsec:multi_calib}, a comprehensive and detailed explanation of the proposed multi-camera hand-eye calibration method is described.


\subsection{Single-camera hand-eye calibration}
\label{subsec:single_calib}
In \cite{evangelista2022unified} we presented an hand-eye calibration technique which does not rely on the PnP algorithm for the estimation of the transformation camera-to-board $A$ (see Figure~\ref{fig:hand_eye_setup}),
but rather solves the calibration through the minimization of the re-projection error.
%

\begin{figure}[t]
  \centering
  \includegraphics[width=0.99\linewidth]{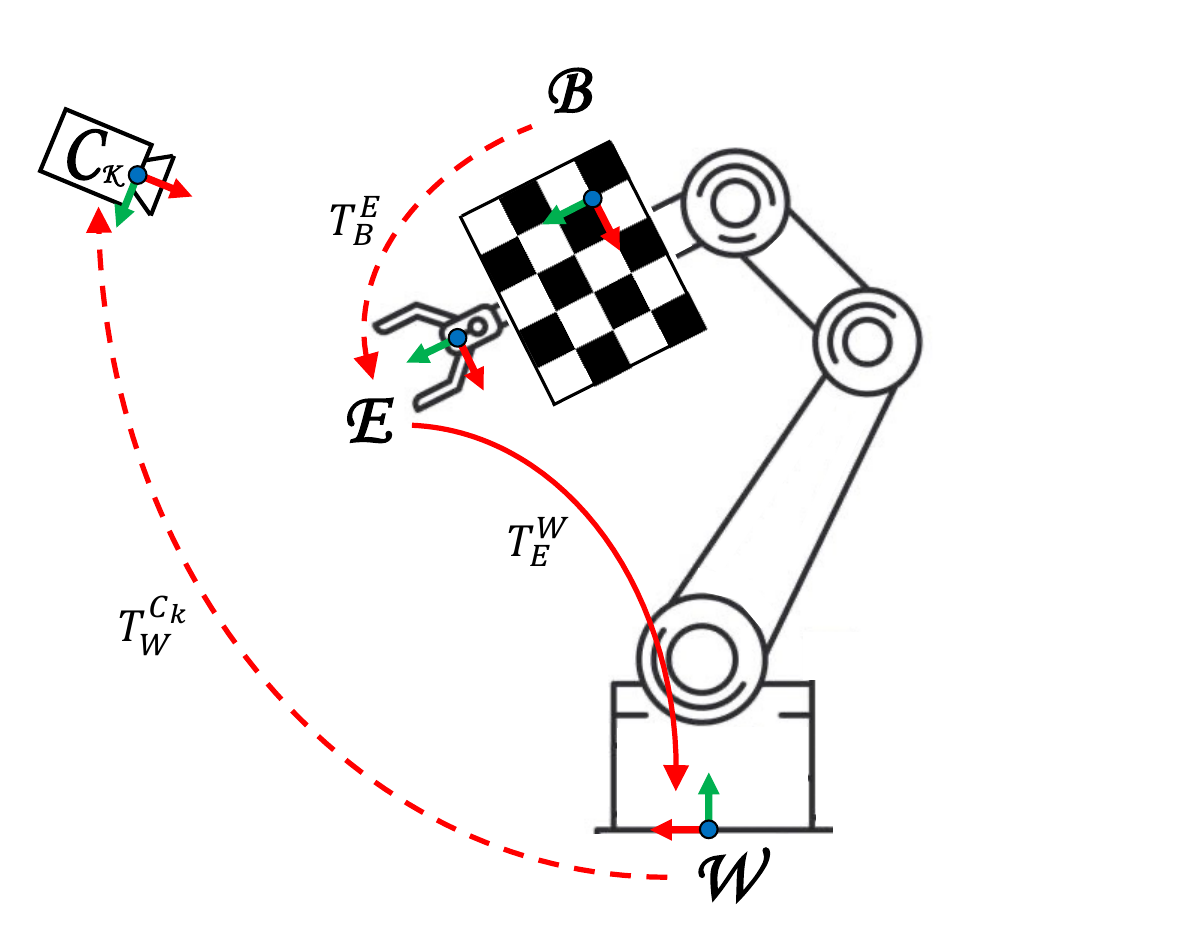}
  \caption{Transformations chain in a single-camera hand-eye setup, illustrating the re-projection of calibration pattern corners onto the image plane. The robot's end effector pose relative to its base (W) is denoted by $T_{E}^{W}$, along with two unknown matrices: $X$, representing the hand-eye transformation, and $Z$, denoting the transformation between the board and the robot's end-effector pose.}
  \label{fig:single_camera_setup}
  \vspace{-0.5cm}
\end{figure}

%
Consider the single-camera setup shown in Figure~\ref{fig:single_camera_setup}. Given a set of $M$ pairs of robot poses $T_E^W$ and images acquired by a camera $C_k$, we aim to estimate the unknown rototranslations $T_W^{C_k}$ and $T_B^E$ by minimizing the objective function $c$ reported in (\ref{eq:cost_function}).
This cost function represents the euclidean distance between the detected 2D corners \mbox{$p_{ij}^{D}=(u_{x}, u_{y})_{ij}^{D}$} of the calibration pattern and their corresponding 3D corners re-projected on the image plane, denoted by $(u_{x}, u_{y})_{ij}^{P}$.
\begin{equation}
     c = \sum_{j=0}^{M-1}\sum_{i=0}^{L-1} \left\|{\begin{pmatrix} u_{x} \\ u_{y} \end{pmatrix}_{ij}^{\mathrm{P}}-\begin{pmatrix} u_{x} \\ u_{y} \end{pmatrix}_{ij}^{\mathrm{D}}}\right\|^{2}
    \label{eq:cost_function}
\end{equation}
Here, $j=0,\dots,M-1$ denotes the $j^{th}$ pose of the robot's end-effector with respect to its base and $i=0,\dots,L-1$ is the $i^{th}$ corner of the calibration pattern.
%

In particular, consider the 3D coordinates $P_{i}^{B}$ of the $i^{th}$ corner in the calibration pattern reference frame $B$, and the function $\pi_{k}(P_{i})$ that describes the projection of a 3D point in the camera frame onto the image plane of a camera $C_k$ with known intrinsic and distortion parameters. The projection of the 3D corners of the calibration pattern onto the image plane is given by:
\begin{equation}
   \begin{pmatrix} u_{x} \\ u_{y} \end{pmatrix}_{ij}^{\mathrm{P}} = \pi_{k}\left(P_{i}^{C_{k}}\right) = \pi_{k}\left(T_{W}^{C_{k}} [T_{E}^{W}]_{j} [T_{B}^{E}]_{k} P_{i}^{B}\right)
    \label{eq:camera_projection}
\end{equation}
where the 3D corners $P_{i}^{B}$ are transformed in the camera frame by means of the chain of transformations depicted in Figure~\ref{fig:single_camera_setup}. While the transformation $\left[T_{E}^W\right]_j$ describing the $j^{th}$ end-effector pose is known from the robot kinematics, the remaining transformations are unknown: the hand-eye transformation $T_{W}^{C_{k}}$, and the rototranslation $[T_{B}^{E}]_{k}$ from the calibration board to the robot's end-effector for camera $C_k$.
Therefore, the cost function for calibrating a single camera $C_k$ can be rewritten as in the equation~(\ref{eq:minimization_eq}). 
\begin{equation}
    c_{k} = \sum_{j=0}^{M-1}\sum_{i=0}^{L-1}\left\|{\pi_{k}\left({T_{W}^{C_{k}}} \left[T_{E}^{W}\right]_{j} [T_{B}^{E}]_{k} P_{i}^{B} \right) -p_{ijk}^{D}}\right\|^{2}
    \label{eq:minimization_eq}
\end{equation}

\subsection{Multi-camera hand-eye calibration}
\label{subsec:multi_calib}
\input{tex/3_multiview}

\section{PERFORMANCE EVALUATION PROCEDURE}
\label{sec:metric}
To comprehensively assess the effectiveness and the robustness of our method, we carried out a series of calibration experiments. These experiments compared our method against other state-of-the-art calibration techniques using the publicly available METRIC dataset and data acquired in two real industrial environments designed for human-robot collaboration. By employing the METRIC dataset, we aimed to validate the calibration method's precision, leveraging the dataset's ground truth data provided for both synthetic and real scenarios. 
Notably, the dataset features images captured by a network of cameras surrounding the robot arm, with three different workcell sizes. This enables a thorough evaluation of our method's performance considering the variations in distances between the cameras and the calibration pattern. 
The experiments conducted in industrial environments aimed to assess our method's suitability and robustness in complex and challenging industrial contexts. These settings are particularly demanding due to their larger robotic workcells and the limited availability of data due to the difficulties associated with  capturing numerous images within such cluttered areas.

In the experiments on METRIC we consider the average translation error ($e^{GT}_{t}$) and rotation error ($e^{GT}_{\theta}$), defined as:
\begin{equation}
\label{eq:error1}
    e^{GT}_{t} = \frac{\sum_{i=0}^{N-1}\left\|t - \hat{t}\right\|_{2}}{N}
\end{equation}

\begin{equation}
\label{eq:error2}
    e^{GT}_{\theta} = \frac{\sum_{i=0}^{N-1}angle(R^{T}\hat{R})}{N}
\end{equation}
where $N$ represents the number of sensors belonging to the camera network. In these equations, $t$ and $R$ represent the translation vector and rotation matrix provided in the ground truth data, while $\hat{t}$ and $\hat{R}$ are the values estimated through the calibration process, all related to the hand-eye transformations $T_{W}^{C_{k}}$. 
Note that rotation error is defined considering the angle of the relative rotation between $R$ and $\hat{R}$ using
the axis-angle representation, which is computed as
$angle()$ in (~\ref{eq:error2}).
%
%
%
For the industrial performance evaluation of our method, since the ground truth data are not available, we adopt the metric used in several hand-eye calibration papers \cite{tabb2017solving}, \cite{wang2022accurate}, \cite{enebuse2021comparative}, which is obtained from the decomposition of the homogeneous equation $AX=ZB$ described in Section~\ref{sec:related_works}. Specifically, the translation error $e_{t}$ and the rotation error $e_{\theta}$ are computed as shown in (\ref{eq:real_metrics_tras}) and (\ref{eq:real_metrics_rot}).
\begin{align}
\label{eq:real_metrics_tras}
e_{t} &= \frac{1}{NM}\sum_{i=0}^{N-1}\sum_{j=0}^{M-1}\left\|(R_{A_{j}}t_{X_{i}} + t_{A_{j}}) - (R_{Z}t_{B_{j}} + t_{Z})\right\|_{2} \\
\label{eq:real_metrics_rot}
e_{\theta} &= \frac{1}{NM}\sum_{i=0}^{N-1}\sum_{j=0}^{M-1} angle\left((R_{A_{j}}R_{X_{i}})^{T}(R_{Z}R_{B_{j}})\right)
\end{align}
Here, $j= 0, \dots, M-1$ denotes the $j^{th}$ robot pose achieved during the image acquisition. The error is evaluated as the average over all the $M$ captured images for all the $N$ cameras. 
In scenarios with no ground truth, as illustrated in the industrial settings discussed in Section \ref{sec:industrial}, only the equations related to the second error metric were employed for evaluation. Instead, when experiments were conducted on the METRIC dataset, the primary metrics for assessment were (\ref{eq:error1}) and (\ref{eq:error2}), with the additional metrics (\ref{eq:real_metrics_tras}) and (\ref{eq:real_metrics_rot}) used to demonstrate the correctness of those metrics and its consistency with the ground truth results. Moreover, for each experiment, we computed the runtime of each hand-eye calibration method used to calibrate the robotic workcell. 


\section{RESULTS ON METRIC DATASET}
\label{sec:experiments}
In this section we report the experimental results obtained on the METRIC dataset~\cite{allegro2023metric}, considering the error metrics defined in Section \ref{sec:metric}. Subsections \ref{sec:synth} and \ref{sec:real} analyze the performance achieved with synthetic and real images from the dataset, respectively. 
Note that the METRIC dataset consider different workcell sizes and an A4 paper checkerboard as calibration pattern, whose inner corners are arranged in a $4\times3$ grid with a spacing of about 5\,cm. 
This allows to investigate how limited sizes of calibration patterns affect calibration performances, which is one of the main limitation of calibrating cameras in large robotic workcells designed for human-robot collaboration tasks.

\subsection{METRIC: SYNTHETIC DATA}
\label{sec:synth}
The synthetic data used in METRIC comprises images obtained within simulated robotic workcells 
with various sizes, encompassing small, medium, and large workcells, covering an area of approximately 6\,m$^2$, 12\,m$^2$, and 20\,m$^2$ respectively.
In Table~\ref{tab:synthetic_evaluation} the errors and the time required for running each algorithm are reported, distinguishing the single-camera and multi-camera methods.
\begin{table*}[t]
\caption{Average errors for all cameras achieved by hand-eye calibration techniques in METRIC simulated workcells. Bold values indicate the top-performing calibration method for each metric.}
\label{tab:synthetic_evaluation}
\centering
\resizebox{\textwidth}{!}{
\begin{tabular}{|c||c|c||c|c||c||c|c||c|c||c||c|c||c|c||c|}
\hline
\multirow{3}{*}{Method} & \multicolumn{5}{c||}{\textbf{Small workcell}} & \multicolumn{5}{c||}{\textbf{Medium workcell}} & \multicolumn{5}{c|}{\textbf{Large workcell}}\\
\cline{2-16}
& \multicolumn{2}{c||}{Ground truth} & \multicolumn{2}{c||}{AX=ZB} & \multirow{2}{*}{Time [s]} & \multicolumn{2}{c||}{Ground truth} & \multicolumn{2}{c||}{AX=ZB} & \multirow{2}{*}{Time [s]} & \multicolumn{2}{c||}{Ground truth} & \multicolumn{2}{c||}{AX=ZB} & \multirow{2}{*}{Time [s]}\\
                        & $e^{GT}_{t}$ [mm] & $e^{GT}_{\theta}$ [deg] & $e_{t}$ [mm] & $e_{\theta}$ [deg] & & $e^{GT}_{t}$ [mm] & $e^{GT}_{\theta}$ [deg] & $e_{t}$ [mm] & $e_{\theta}$ [deg] & & $e^{GT}_{t}$ [mm] & $e^{GT}_{\theta}$ [deg] & $e_{t}$ [mm] & $e_{\theta}$ [deg] & \\
\hline
Tsai~\cite{tsai1989new} & $427.16 $ & $2.06$ & $40.26$ & $0.62$ & $0.08$ & $820.904$ & $6.72$ & $47.29$ & $1.12$ & $0.11$ & $587.59$ & $0.21$ & $35.47$ & $0.56$ & $0.095$\\
Park~\cite{park1994robot} & $3.27$ & $0.04$ & $2.35$ & $0.12$ & $0.123$ & $5.27$ & $0.03$ & $4.47$ & $0.18$ & $0.16$ & $12.06$ & $0.20$ & $4.97$ & $0.18$ & $0.133$ \\
Shah~\cite{shah2013solving}  & $1.90$ & $0.05$ & $2.63$ & $0.13$ & $0.07$ & $3.03$ & $0.05$ & $4.37$ & $0.19$ & \pmb{$0.08$} & $5.35$ & $0.08$ & $4.51$ & $0.16$ & $0.12$\\
Danililidis~\cite{daniilidis1996dual} & $42.43$ & $0.58$ & $15.34$ & $0.37$ & $0.122$ & $100.09$ & $3.27$ & $39.24$ & $1.08$ & $0.166$ & $57.58$ & $0.17$ & $24.9$ & $0.7$ & $0.126$\\
Andreff~\cite{andreff1999line}& $10.05$ & $0.05$ & $8.72$ & $0.33$ & $0.262$ & $91.83$ & $0.38$ & $80.84$ & $1.3$ & $0.327$ & $149.53$ & $0.34$ & $116.54$ & $2.89$ & $0.267$\\
Li~\cite{li2010simultaneous} & $1.93$ & $0.05$ & $2.71$ & $0.14$ & \pmb{$0.06$} & $3.15$ & $0.05$ & $4.55$ & $0.19$ & $0.09$ & $5.87$ & $0.08$ & $4.58$ & $0.16$ & \pmb{$0.09$}\\
Evangelista~\cite{evangelista2022unified} & $1.93$ & $0.03$ & $1.98$ & $0.05$ & $14.99$ & $5.46$ & $0.07$ & $6.43$ & $0.08$ & $15.46$ & $19.98$ & $0.29$ & $35.21$ & $0.53$ & $25.1$\\
Koide~\cite{koide2019general} & $1.68$ & $0.03$ & $1.46$ & $0.07$ & $116.1$ & $3.94$ & $0.02$ & $2.31$ & $0.11$ & $88.82$ & $1236.3$ & $0.04$ & $773.11$ & $5.38$ & $77.38$\\
\hline
Evangelista~\cite{evangelista2023graph} & $1.02$ & \pmb{$0.02$} & $1.14$ & $0.05$ & $237.97$ & $3.38$ & \pmb{$0.02$} & $2.02$ & $0.1$ & $311.67$ & $1.43$ & $0.02$ & $4.02$ & $0.15$ & $249.89$\\
Tabb~\cite{tabb2017solving} & $2.52$ & $0.2$ & $2.86$ & $0.32$ & $69.45$ & $5.79$ & $0.36$ & $7.35$ & $0.38$ & $87.92$ & $13.52$ & $0.89$ & $17.45$ & $0.65$ & $87.98$\\
Ours \ref{subsec:multi_calib} & \pmb{$0.71$} & \pmb{$0.02$} & \pmb{$0.45$} & \pmb{$0.03$} & $4.67$ & \pmb{$0.75$} & \pmb{$0.02$} & \pmb{$0.83$} & \pmb{$0.05$} & $13.78$ & \pmb{$1.08$} & \pmb{$0.01$} & \pmb{$2.02$} & \pmb{$0.09$} & $27.13$\\
\hline
\end{tabular}}
\vspace{-0.5cm}
\end{table*}
The results clearly prove that the robotic workcell size has a significant impact on the calibration accuracy for various hand-eye calibration methods. Notably, as the mean distance between sensors and the calibration pattern extends, introducing a more complex scenario for corner detection, consistently both translation errors  ($e_{t}$, $e_{t}^{GT}$) and rotation errors ($e_{\theta}$,  $e_{\theta}^{GT}$) exhibit a decline, in accordance with our previous results~\cite{allegro2023metric}. It is observed that numerous methods~\cite{park1994robot,daniilidis1996dual,evangelista2022unified, tabb2017solving} experience a notable reduction in calibration efficacy within larger robotic workcells, while others~\cite{tsai1989new,andreff1999line,koide2019general} may even diverge from the ideal solution in the large robotic workcell. However, the multi-camera hand-eye calibration method presented in this paper emerges as the most robust and consistently accurate in all three scenarios, demonstrating minimal sensitivity to variations in robotic workcell size across all error metrics.

In particular, in the case of the large robotic workcell, the proposed method achieves remarkable results, ensuring translation errors of approximately 1\,mm with respect to the ground truth, and rotation errors as low as 0.01\,deg. The method's robustness to potential misdetections in large workcells is due to its optimization process implementation, which not only focuses on minimizing re-projection errors on a single camera but also leverages minimization across other cameras, particularly when they simultaneously capture the calibration pattern.
Conversely, as expected in terms of optimization time, single-camera calibration methods (i.e., first group of rows in Table~\ref{tab:synthetic_evaluation}) prove significantly faster, benefiting from solving smaller sets of homogeneous equations and generally exhibiting lower algorithmic complexity. However, among multi-camera calibration methods, the presented approach stands out as the fastest, offering a balance between accuracy and efficiency.

\subsection{METRIC: REAL DATA}
\label{sec:real}
  
The real images of METRIC were captured in two robotic workcells of different size—one small, covering an area of about 7\,m$^{2}$ typically designed for tasks that involve transferring small objects between human operators and robots in minor assembly applications, and the large one, approximately spanning an area of 15 m$^{2}$, configured for human-robot collaboration applications involving several people within the workcell, e.g. for the collaborative transport of large objects.
For each robotic workcell layout, three different sets of images were collected by means of different camera networks, each characterized by the use of a particular type of sensor: Intel RealSense Lidar camera L515, Intel RealSense Depth D455 sensor and the Microsoft Kinect V2.
As discussed in \cite{allegro2023metric}, the size of the workcell is not the only factor influencing calibration; the type of sensor and its characteristics also play a significant role in the calibration pattern detection and, consequently, in the calibration process.
\begin{table*}[t]
\caption{Average error achieved by the hand-eye calibration techniques in the small-size and large-size real workcells of METRIC. Bold results indicate the top-performing calibration method for each metric (“$-$” denotes  the non-convergence).}
\label{tab:real_evaluation}
\centering
\resizebox{\textwidth}{!}{
\begin{tabular}{|c||c|c||c|c||c||c|c||c|c||c||c|c||c|c||c|}
\hline
\multirow{4}{*}{Method} & \multicolumn{15}{c||}{\textbf{Small real workcell}}\\ \cline{2-16} & \multicolumn{5}{c||}{\textbf{Microsoft Kinect V2}} & \multicolumn{5}{c||}{\textbf{Intel RealSense Depth D455}} & \multicolumn{5}{c|}{\textbf{Intel RealSense LiDAR L515}}\\
\cline{2-16}
& \multicolumn{2}{c||}{Ground truth} & \multicolumn{2}{c||}{AX=ZB} & \multirow{2}{*}{Time [s]} & \multicolumn{2}{c||}{Ground truth} & \multicolumn{2}{c||}{AX=ZB} & \multirow{2}{*}{Time [s]} & \multicolumn{2}{c||}{Ground truth} & \multicolumn{2}{c||}{AX=ZB} & \multirow{2}{*}{Time [s]}\\
                        & $e^{GT}_{t}$ [mm] & $e^{GT}_{\theta}$ [deg] & $e_{t}$ [mm] & $e_{\theta}$ [deg] & & $e^{GT}_{t}$ [mm] & $e^{GT}_{\theta}$ [deg] & $e_{t}$ [mm] & $e_{\theta}$ [deg] & & $e^{GT}_{t}$ [mm] & $e^{GT}_{\theta}$ [deg] & $e_{t}$ [mm] & $e_{\theta}$ [deg] & \\
\hline
Tsai~\cite{tsai1989new} & $75.13$ & $0.15$ & $12.11$
 & $0.90$ & \pmb{$0.07$} & $1215.45$ & $14.32$ & $224.68$ & $10.65$ & $0.09$ & $396.83$ & $0.59$ & $53.51$ & \pmb{$0.86$} & \pmb{$0.10$}\\
Park~\cite{park1994robot} & $56.61$ & $0.14$ & $10.42$ & $0.86$ & $0.11$ & $167.70$ & $3.48$ & $69.20$ & $2.13$ & $0.08$ & $72.40$ & $0.12$ & $40.32$ & $1.79$ & $0.13$ \\
Shah~\cite{shah2013solving}  & $27.22$ & $0.75$ & $10.37$ & $0.94$ & $0.08$ & $23.90$ & $0.54$ & $65.90$ & $2.31$ & $0.08$ & $18.51$ & $0.34$ & $30.93$ & $1.68$ & $0.12$\\
Danililidis~\cite{daniilidis1996dual} & $49.62$ & $0.14$ & $10.32$ & $0.94$ & $0.11$ & $1425.42$ & $0.43$ & $541.38$ & $1.69$ & $0.09$ & $316.34$ & $0.30$ & $71.11$ & $1.77$ & $0.16$\\
Andreff~\cite{andreff1999line}& $235.17$ & $0.58$ & $127.61$ & $5.51$ & $0.22$ & $1101.90$ & $11.95$ & $759.66$ & $11.76$ & $0.19$ & $596.57$ & $6.78$ & $443.67$ & $4.75$ & $0.27$\\
Li~\cite{li2010simultaneous} & $66.67$ & $0.73$ & $13.89$ & $0.93$ & $0.10$ & $51.03$ & $0.72$ & $100.29$ & $2.40$ & \pmb{$0.07$} & $23.70$ & $0.31$ & $34.27$ & $1.64$ & $0.11$\\
Evangelista~\cite{evangelista2022unified} & $42.79$ & $0.57$ & $13.21$ & $0.81$ & $7.54$ & $45.14$ & $0.43$ & $72.91$ & $1.22$ & $13.89$ & $26.20$ & $0.39$ & $50.33$ & $3.58$ & $31.27$\\
Koide~\cite{koide2019general} & $46.52$ & $0.12$ & $12.84$ & $0.68$ & $125.14$ & $72.87$ & $0.28$ & $22.30$ & $2.91$ & $53.30$ & $36.67$ & $0.78$ & $82.95$ & $2.96$ & $122.62$\\
\hline
Evangelista~\cite{evangelista2023graph} & $42.42$ & \pmb{$0.09$} & $7.31$ & $0.56$ & $215.23$ & $41.80$ & $0.34$ & $11.10$ & $1.13$ & $210.61$ & $24.11$ & $0.44$ & $33.75$ & $1.70$ & $198.42$\\
Tabb~\cite{tabb2017solving} & $51.57$ & $0.73$ & $11.20$ & $0.81$ & $145.31$ & $63.98$ & $1.36$ & $81.32$ & $2.76$ & $65.22$ & $34.59
$ & $0.94$ & $67.23$ & $3.56$ & $89.21$\\
Ours \ref{subsec:multi_calib} & \pmb{$22.01$} & \pmb{$0.09$} & \pmb{$6.54$} & \pmb{$0.55$} & $16.79$ & \pmb{$13.21$} & \pmb{$0.07$} & \pmb{$9.21$} & \pmb{$0.66$} & $20.11$ & \pmb{$13.68$} & \pmb{$0.02$} & \pmb{$24.95$} & $1.32$ & $42.21$\\
\hline
\hline
\multirow{4}{*}{Method} & \multicolumn{15}{c||}{\textbf{Large real workcell}}\\ \cline{2-16} & \multicolumn{5}{c||}{\textbf{Microsoft Kinect V2}} & \multicolumn{5}{c||}{\textbf{Intel RealSense Depth D455}} & \multicolumn{5}{c|}{\textbf{Intel RealSense LiDAR L515}}\\
\cline{2-16}
& \multicolumn{2}{c||}{Ground truth} & \multicolumn{2}{c||}{AX=ZB} & \multirow{2}{*}{Time [s]} & \multicolumn{2}{c||}{Ground truth} & \multicolumn{2}{c||}{AX=ZB} & \multirow{2}{*}{Time [s]} & \multicolumn{2}{c||}{Ground truth} & \multicolumn{2}{c||}{AX=ZB} & \multirow{2}{*}{Time [s]}\\
                        & $e^{GT}_{t}$ [mm] & $e^{GT}_{\theta}$ [deg] & $e_{t}$ [mm] & $e_{\theta}$ [deg] & & $e^{GT}_{t}$ [mm] & $e^{GT}_{\theta}$ [deg] & $e_{t}$ [mm] & $e_{\theta}$ [deg] & & $e^{GT}_{t}$ [mm] & $e^{GT}_{\theta}$ [deg] & $e_{t}$ [mm] & $e_{\theta}$ [deg] & \\
\hline
Tsai~\cite{tsai1989new} & $1924.88$ & $18.78$ & $318.14$ & $15.30$ & \pmb{$0.09$} & $2759.10$ & $14.91$ & $461.62$ & $15.25$ & \pmb{$0.06$} & $1881.75$ & $11.53$ & $222.23$ & $7.50$ & \pmb{$0.08$}\\
Park~\cite{park1994robot} & $336.86$ & $1.08$ & $178.91$ & $4.69$ & $0.14$ & $355.10$ & $5.33$ & $226.85$ & $5.10$ & $0.09$ & $233.57$ & $3.98$ & $70.63$ & $1.60$ & $0.13$ \\
Shah~\cite{shah2013solving}  & $54.92$ & $0.73$ & $75.34$ & $2.62$ & $0.12$ & $-$ & $-$ & $-$ & $-$ & $-$ & $26.15$ & $0.31$ & $38.12$ & $2.02$ & $0.09$\\
Danililidis~\cite{daniilidis1996dual} & $1757.97$ & $13.23$ & $629.15$ & $6.07$ & $0.14$ & $-$ & $-$ & $-$ & $-$ & $-$ & $8359.06$ & $0.32$ & $4176.45$ & $1.60$ & $0.13$\\
Andreff~\cite{andreff1999line}& $2370.61$ & $12.34$ & $1639.04$ & $12.20$ & $0.29$ & $2554.29$ & $35.89$ & $1680.34$ & $7.25$ & $0.18$ & $1881.23$ & $12.79$ & $1310.25$ & $10.94$ & $0.26$\\
Li~\cite{li2010simultaneous} & $129.39$ & $0.72$ & $111.44$ & $2.61$ & $0.13$ & $-$ & $-$ & $-$ & $-$ & $-$ & $26.39$ & $0.30$ & $66.42$ & $2.49$ & $0.12$\\
Evangelista~\cite{evangelista2022unified} & $77.26$ & $0.77$ & $123.12$ & $3.21$ & $58.32$ & $136.39$ & $1.33$ & $57.32$ & $4.32$ & $14.32$ & $60.59$ & $0.43$ & $54.36$ & $1.66$ & $14.37$\\
Koide~\cite{koide2019general} & $69.38$ & $0.35$ & $20.72$ & $1.04$ & $201.45$ & $65.39$ & $0.32$ & $45.21$ & $2.13$ & $28.41$ & $59.58$ & $0.11$ & $15.05$ & $1.05$ & $25.86$\\
\hline
Evangelista~\cite{evangelista2023graph} & $63.54$ & \pmb{$0.24$} & $15.79$ & $0.71$ & $275.87$ & $59.33$ & $0.22$ & $48.50$ & \pmb{$2.10$} & $105.91$ & $50.18$ & $0.09$ & \pmb{$11.63$} & $0.89$ & $260.59$\\
Tabb~\cite{tabb2017solving} & $105.01$ & $0.75$ & $154.21$ & $5.32$ & $165.22$ & $153.20$ & $1.74$ & $167.34$ & $6.35$ & $78.23$ & $75.33$ & $0.77$ & $45.23$ & $1.21$ & $78.23$\\
Ours \ref{subsec:multi_calib} & \pmb{$51.18$} & $0.30$ & \pmb{$12.17$} & \pmb{$0.65$} & $77.99$ & \pmb{$45.18$} & \pmb{$0.16$} & \pmb{$35.98$} & $2.12$ & $35.43$ & \pmb{$19.34$} & \pmb{$0.09$} & $12.45$ & \pmb{$0.85$} & $17.33$\\
\hline
\end{tabular}}
\vspace{-0.3cm}
\end{table*}
As illustrated in Table~\ref{tab:real_evaluation}, the multi-camera hand-eye calibration method proposed in this work consistently outperforms other methods and it ensures convergence to an optimal solution, not achievable by some single-camera methods~\cite{shah2013solving, li2010simultaneous, daniilidis1996dual}. This can primarily be attributed to the incorporation of two additional constraints, which enhance the algorithm's robustness in real-world scenarios where corner detection may be imprecise. In particular, our method achieves a translation error $e_{t}^{GT}$ lower than 52\,mm and a rotation error $e_{\theta}^{GT}$ lower than 0.3\,deg,  demonstrating effectiveness even within the larger workcell.
Notably, even with the lower-resolution Intel RealSense D455 sensor, the proposed method achieves an average error comparable to that obtained with the other sensors, guaranteeing superior performance with respect to all other state-of-the-art methods.
In general, multi-camera calibration methods demonstrate greater accuracy compared to running N times the single-camera methods, which do not have the possibility of mitigating the impact of inaccurate pattern detection through other points of view that are simultaneously considered.
%

Our optimization process, while not as rapid as single-camera methods that address calibration problem through closed-form solutions \cite{tsai1989new,park1994robot, daniilidis1996dual,andreff1999line,shah2013solving,li2010simultaneous}---which complete in less than 1\,second--- is comparable with single-camera approaches~\cite{evangelista2022unified,koide2019general} that employ non-linear and graph optimization techniques. Moreover, it stands out as the quickest among other multi-camera methods, achieving calibration in some instances over ten times faster. It's important to note that while some methods may achieve faster results, they often do so at the cost of accuracy. Our approach, however, maintains an optimal balance, offering both speed and precision, thereby positioning it among the top-performing strategies.
%
The availability of ground truth on real data in METRIC makes it possible to verify the reliability of the metrics (\ref{eq:real_metrics_tras}) and (\ref{eq:real_metrics_rot}) in a real-world scenario, as shown in Table~\ref{tab:real_evaluation} where they follows the same trend of GT-based metrics (\ref{eq:error1}) and (\ref{eq:error2}) for the various methods tested.

\section{RESULTS ON INDUSTRIAL SCENARIOS}
\label{sec:industrial}
\begin{figure}[t]
  \centering
  \includegraphics[width=\linewidth]{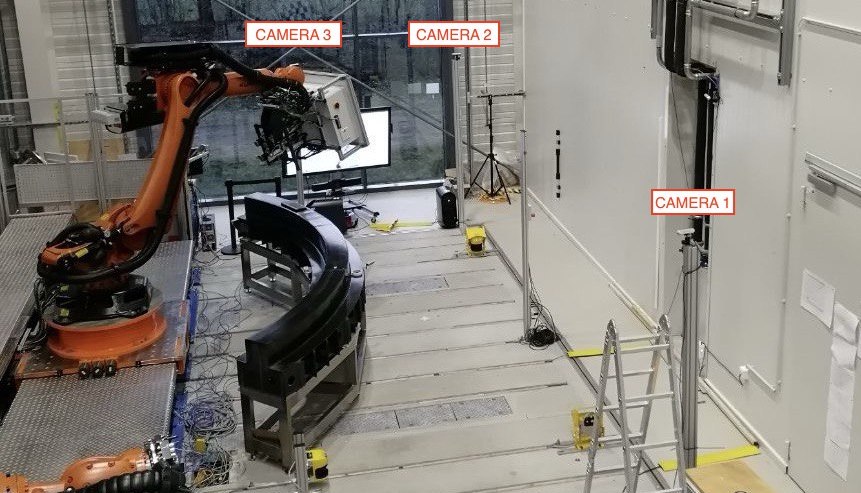}
  \caption{Kuka industrial workcell equipped with three Intel RealSense Depth D455 surrounding a Kuka manipulator at an average distance of 6 meters.}
  
  \label{fig:real_workcell2}
  \vspace{-0.4cm}
\end{figure}

To comprehensively evaluate the robustness of our calibration method even in real industrial environments, we performed calibration experiments in two industrial robotic workcells developed as use cases in the DrapeBot project~\cite{terreran2022smart}. Both workcells have been designed for human-robot collaboration and collaborative transportation of flexible materials, covering an area of about 20~m$^{2}$.  
The former, illustrated in Figure \ref{fig:robotic_workcell} and identified as the ABB robotic workcell, was equipped with four Intel RealSense Depth D455 sensors surrounding an ABB industrial manipulator at an average distance of 4 meters from the base of the robot. 
The second workcell, designated as the Kuka industrial workcell, was equipped with a Kuka industrial manipulator surrounded by three RealSense Depth D455 sensors at an average distance of 6 meters from the robot base (Figure \ref{fig:real_workcell2}). 
Within these two robotic workcells, the image collection comprised fewer than 10 images for the first workcell and exactly 15 for the second. This limited number was attributed to the challenges associated with moving the robot to various locations and orientations in front of the sensors, due to the presence of large objects, such as the mold depicted in Figure~\ref{fig:workcell_calib} and Figure~\ref{fig:real_workcell2}, near the robot, which hinders its movement. The calibration pattern was composed of a checkerboard with $4\times3$ inner corners, spaced approximately 6\,cm apart. The results of the calibration process are reported in Table \ref{tab:industrial_workcells}. These results are assessed using the metrics defined in (\ref{eq:real_metrics_tras}) and (\ref{eq:real_metrics_rot}), due to the absence of ground truth data.
\begin{table}[h]
\caption{Average error achieved by the hand-eye calibration techniques in ABB and Kuka industrial workcells.}
\label{tab:industrial_workcells}
\centering
\resizebox{\columnwidth}{!}{
\begin{tabular}{|c||c|c||c||c|c||c|}
\hline
\multirow{3}{*}{Method} & \multicolumn{3}{c||}{\textbf{ABB industrial workcell}} & \multicolumn{3}{c|}{\textbf{Kuka industrial workcell}} \\
\cline{2-7}
& \multicolumn{2}{c||}{AX=ZB} & \multirow{2}{*}{Time [s]} & \multicolumn{2}{c||}{AX=ZB} & \multirow{2}{*}{Time [s]}\\
                        & $e_{t}$ [mm] & $e_{\theta}$ [deg] & &$e_{t}$ [mm] & $e_{\theta}$ [deg] &   \\
\hline
Tsai~\cite{tsai1989new} & $172.07$ & $10.15$ & \pmb{$0.09$} & $30.28$ & $1.27$ & $0.13$ \\
Park~\cite{park1994robot} & $83.54$ & $11.93$ & $0.12$ & $27.82$ & $2.21$ & $0.17$ \\
Shah~\cite{shah2013solving}& $106.99$ & $10.76$ & $0.10$ & $72.94$ & $1.52$ & $0.15$ \\
Danililidis~\cite{daniilidis1996dual}& $77.09$ & $11.14$ & $0.13$ & $36.33$ & $1.42$ & $0.14$ \\
Andreff~\cite{andreff1999line}& $1378.25$ & $6.06$ & $0.12$ & $71.37$ & $1.66$ & $0.21$ \\
Li~\cite{li2010simultaneous}& $-$ & $-$ & $-$ & $652.44$ & $1.23$ & \pmb{$0.11$} \\
Evangelista~\cite{evangelista2022unified}& $718.94$ & $7.52$ & $6.32$ & $485.92$ & $6.15$ & $12.97$ \\
Koide~\cite{koide2019general} & $131.08$ & $9.60$ & $7.12$ & $1737.68$ & $10.23$ & $9.45$ \\
\hline
Evangelista~\cite{evangelista2023graph} & $11.95$ & $2.65$ & $11.32$ & $17.38$ & $0.78$ & $36.78$ \\
Tabb~\cite{tabb2017solving} & $53.76$ & $3.18$ & $8.98$ & $22.65$ & $1.23$ & $17.67$ \\
Ours \ref{subsec:multi_calib} & \pmb{$6.31$} & \pmb{$0.98$} & $3.12$ & \pmb{$14.12$} & \pmb{$0.67$} & $7.15$ \\
\hline
\end{tabular}}
\vspace{-0.4cm}
\end{table}

Table \ref{tab:industrial_workcells} highlights that our proposed multi-camera hand-eye calibration method achieves high accuracy with a translation error of approximately $1$\,cm and a rotation error lower than $1$\,deg. In contrast, some single-camera calibration methods (e.g.,~\cite{li2010simultaneous,evangelista2022unified,koide2019general}) struggle to converge to an optimal solution, proving to be unsuitable for such challenging scenarios. Notably, our method shows exceptional accuracy with a minimum number of images, less than 10 for the ABB robotic workcell and less than 15 in the Kuka robotic workcell. 
This efficiency is significant, especially in the context of industrial robotic workcells, where acquiring images is challenging due to obstacles that restrict flexible robot movement. Consequently, this approach significantly reduces calibration times and minimizes interruptions in production lines.

\section{CONCLUSIONS}
\label{sec:conclusions}

This work proposed a multi-camera hand-eye calibration, incorporating in the optimization process two additional constraints not previously considered together in the existing literature, namely the common board-to-end-effector transformation across all cameras and relative camera-to-camera transformations among all sensors. The proposed method was evaluated on the publicly available METRIC dataset, allowing for a thorough assessment of calibration accuracy and robustness in robotic workcells of various sizes equipped with different sensors. Through a comprehensive analysis, our method significantly outperformed other state-of-the-art calibration methods, showcasing an excellent balance between speed and calibration accuracy. 

The proposed method was also tested in two real-world industrial scenarios, considering robotic workcells designed for human-robot collaboration: it provided the highest accuracy also in such scenarios, resulting once again the most robust approach. Notably, despite the reduced number of images for calibration, it achieved outstanding results compared to other state-of-the-art methods, minimizing execution time.
In light of these results, we have made the calibration method publicly available, and a potential future direction could involve extending this method to a multi-robot and multi-camera scenario.






\section*{ACKNOWLEDGMENT}
The research leading to these results has received funding from the European Union's Horizon 2020 research and innovation program under grant agreement No. 101006732 (DrapeBot).

\bibliographystyle{IEEEtran}
\bibliography{references}

\end{document}

%% file: tex/3_multiview.tex
%

The method presented in Section~\ref{subsec:single_calib} could easily be used to calibrate a network of cameras by simply applying it separately to each camera in the network. However, in the case of a network of $N$ cameras this approach leads to estimating $N$ hand-eye transformations $T_W^{C_k}$ and $N$ board-to-end-effector transformations $\left[T_B^E\right]_k$ independent of each other. On one hand, this can be very inefficient since $N$ different calibration processes are needed; on the other hand, this can leads to poor calibration performance of entire camera network since the relative transformation $T_{C_{t}}^{C_{k}}$ between cameras can be affected by accumulated errors in translation and rotation by chaining their corresponding hand-eye calibration $T_W^{C_{k}}$, $T_W^{C_{t}}$.  
This motivates us to introduce two main types of constraints in the multi-camera calibration process to better exploit the data available in such a scenario, namely: (i) a common single board-to-end-effector transformation $T_{B}^{E}$ for all cameras and (ii) the relative camera-to-camera transformation $T_{C_{t}}^{C_{k}}$ for each pair of cameras $(C_t, C_k)$ which detect the calibration pattern at the same acquisition step.

The former constraint derived from the fact that all cameras are calibrated using the same pattern rigidly attached to the robot, and all images are acquired simultaneously moving just the robot arm: the estimated board-to-end-effector transformation should then be the same for all cameras in the network.
Generalizing (\ref{eq:minimization_eq}) to a multi-camera setup, the overall cost function to be minimized is given by the sum of the re-projection errors of the $N$ cameras, imposing the same transformation $T_{B}^{E}$ for all cameras:
\begin{equation}
    c_{rpj} = \sum_{j=0}^{M-1}\sum_{k=0}^{N-1}\sum_{i=0}^{L-1} \left\| \pi_{k}(T_{W}^{C_{k}}[T_{E}^{W}]_{j}T_{B}^{E}) P_{i}^{B} - p_{ijk}^{D} \right\|^{2}
    \label{eq:c1}
\end{equation}

\begin{figure}
  \centering
  \includegraphics[width=1\linewidth]{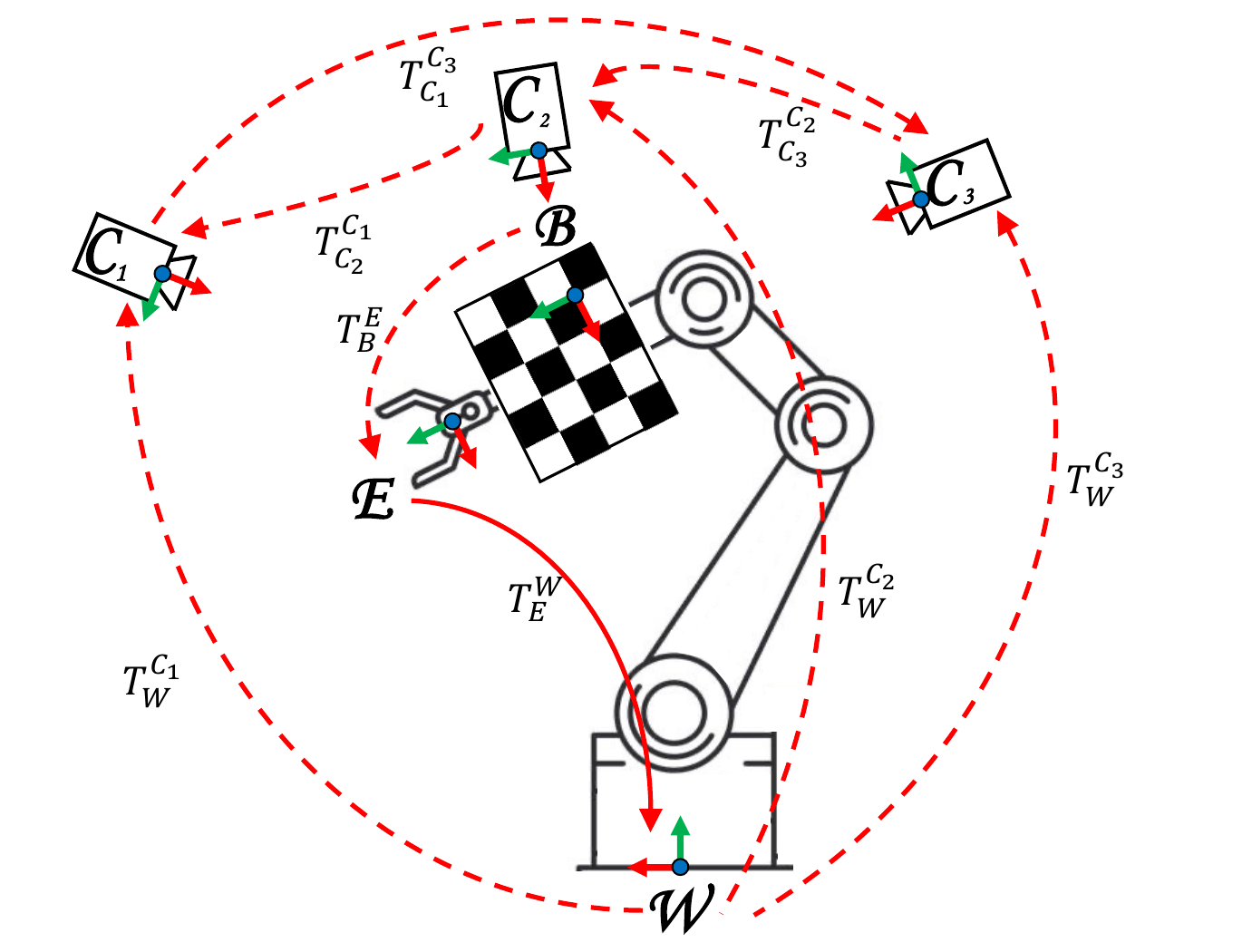}
  \caption{Multi-camera hand-eye setup, illustrating the geometric transformations optimized through the proposed multi-camera hand-eye calibration method. They include the single board-to-end-effector transformation common to all cameras, the spatial constraints among the cameras, and all the hand-eye transformations.}
  \label{fig:multi_camera_setup}
  \vspace{-0.4cm}
\end{figure}

Generally, cameras are positioned to minimize
occlusions and capture different viewpoints of the same
scene simultaneously. The same calibration pattern can thus be observed by multiple cameras simultaneously during the data collection phase. As shown in Figure~\ref{fig:multi_camera_setup}, this introduces an additional path to project the pattern's 3D corners onto the image plane of camera $C_k$: we can either use the transformation $T_W^{C_k}$ as in (\ref{eq:camera_projection}) or the transformation $T_{C_{t}}^{C_{k}}T_{W}^{C_{t}}$ passing through a camera $C_{t}$ which concurrently detected the pattern.

Based on such observation, we proposed an additional cost function $c_{cross}$ which aims to minimize the difference between corners detected by camera $C_k$ and their re-projection onto this camera's image plane through camera $C_t$:
\begin{equation}
    c_{cross} = \sum_{j=0}^{M-1}\sum_{k=0}^{N-1}\sum_{t=0}^{N-1}\sum_{i=0}^{L-1} \left\| p_{ijkt}^{cross} - p_{ijk}^{D} \right\|^{2}
    \label{eq:c2}
\end{equation}
This process is crucial for refining the relative transformations $T_{C_{t}}^{C_{k}}$ between the cameras, exploiting the occurrence of cross-detections---when the checkerboard is simultaneously detected by more cameras. 
A cross-detection matrix $\mathbf{X}_j$ is used within the optimization framework to describe when two cameras are jointly detecting the calibration pattern. This matrix consists of binary values in each cell, $\mathbf{X}_j(k, t)$, where $k\neq t$, indicating the concurrent detection of the calibration pattern by cameras $k$ and $t$ at the $j^{th}$ time step. Consequently, the re-projected corners $p_{ijkt}^{cross}$ can be computed as shown in (\ref{eq:cross_proj}).
\begin{equation}
    p_{ijkt}^{cross} = \pi_{k}\left(T_{C_{t}}^{C_{k}}T_{W}^{C_{t}}[T_{E}^{W}]_{j}T_{B}^{E}P_{i}^{B}\right) \cdot\mathbf{X}_{j}(k,t)
    \label{eq:cross_proj}
\end{equation}

Overall, the cost function to be optimized can be summarized as the sum of two main contributions, the term $c_{rpj}$ aiming to minimize the re-projection error for each individual camera and the term $c_{cross}$ to impose constraints on the relative pose of pairs of cameras:
\begin{equation}
    \underset{T_{C_{t}}^{C_{k}}, T_{W}^{C_{t}},T_{B}^{E}}{\mathrm{argmin}} \, c_{rpj} + c_{cross}
    \label{eq:final_optimization}
\end{equation}
To our knowledge, this is the first multi-camera hand-eye calibration that takes advantage of both constraints through a non-linear optimization algorithm. The proposed method is implemented using the Ceres Solver~\cite{Agarwal_Ceres_Solver_2022} library, adding a Cauchy loss function to enhance resilience against outliers.